# Regional prediction of hematoma expansion from non-contrast computed tomography in spontaneous intracerebral hemorrhage patients

Natasha Ironside, MBChB[1]; Hongyu Kan, MS[17]; Kristofor Pas, BS[2]; Mohammad Shifat-E-Rabbi, PhD[3]; Benjamin Newman, PhD[4]; Shinjini Kundu, MD, PhD[5]; Kareem El Naamani, MD[6]; Tanvir Rizvi, MD[4]; Andrea Becerril-Gaitan, MD[7]; Harrison Snyder, MD[8], Aarathi Manchikalapudi, BSc[9]; Sarah Ratcliffe, PhD[10]; Stacie L. Demel, DO, PhD[11]; Carl D. Langefeld, PhD[12]; Daniel Woo, MD[13]; Stephan Mayer, MD[14]; E. Sander Connolly, MD[15]; Gustavo Rohde, PhD[2, 16,17] for the ERICH Investigators and the VISTA-ICH collaboration*

[1]*Department of Neurological Surgery, University of Illinois Chicago, Chicago, Il*
[2]*Department of Biomedical Engineering, University of Virginia Health System, Charlottesville, VA*
[3]*Department of Electrical and Computer Engineering, North South University, Dhaka, Bangladesh*
[4]*Department of Radiology, University of Virginia Health System, Charlottesville, VA*
[5]*Department of Radiology, Washington University in St. Louis, St. Louis, MO*
[6]*Department of Neurological Surgery, University of Arizona College of Medicine, Tucson, AZ*
[7]*Department of Neurological Surgery, University of Louisville, Louisville, KY*
[8]*Department of Neurological Surgery, Tufts University Medical Center, Boston, MA*
[9]*Department of Neurological Surgery, University of Virginia Health System, Charlottesville, VA*
[10]*Department of Biostatistics, University of Virginia, Charlottesville, VA*
[11]*Department of Neurology and Rehabilitation Medicine, University of Cincinnati, Cincinnati, OH*
[12]*Department of Biostatistics and Data Science, Wake Forest University School of Medicine, Winston-Salem, NC*
[13]*Department of Neurology, University of Buffalo, Buffalo, NY*
[14]*Department of Neurology, Westchester Medical Center, Westchester, NY*
[15] *Department of Neurological Surgery, Vagelos College of Physicians and Surgeons, Columbia University, New York, NY*
[16] *Department of Electrical and Computer Engineering, University of Virginia, Charlottesville, VA*
[17] *Department of Electrical and Computer Engineering, George Washington University, Washington, DC*

***VISTA-ICH Steering Committee:** DF Hanley (Chair), K Butcher, S Davis, B Gregson, KR Lees, P Lyden, S Mayer, K Muir, and T Steiner

**Running title:** Regional hematoma expansion prediction after ICH

**Figures/Tables:** 6

**Funding:** This research was supported by NREF Young Clinician Investigator Award and the National Institutes of Health R01GM130825-01

**Correspondence:**

Natasha Ironside, MBChB
Department of Neurological Surgery
University of Illinois, Chicago
912 S. Wood St.
Chicago
Il 60612
USA
natashai@uic.edu

## Abstract

### Background

The combination of hematoma location, expansion and final volume influence neurological outcome. The aim of this study was to derive and validate a quantitative and interpretable region-based NCCT image feature model for predicting hematoma expansion.

### Methods

The Ethnic/Racial Variations of Intracerebral Hemorrhage(ERICH) study was used for derivation and the Virtual International Trials in Stroke Archive(VISTA) was used for validation. NCCT images at presentation and 24$\pm$6hours were pre-processed, segmented and categorized into lobar, basal ganglia or thalamic regions. Hematoma expansion was defined by region and according to its effect on clinical outcome. Transport-based morphometry (TBM) produced a unique transport-based representation from each NCCT hematoma image. Linear classifiers applied to transport representations were used to derive a predictive model. The regional TBM model's performance was compared to alternate methods using AUROC analyses.

### Results

The derivation dataset included 574 patients, comprising 162 lobar, 257 basal ganglia and 155 thalamic hematomas. The validation dataset included 196 patients, comprising 31 lobar, 119 basal ganglia and 46 thalamic hematomas. A combined regional model predicted hematoma expansion in the validation dataset with an AUROC of 0.787. The principal image features which varied by location were size, heterogeneity, eccentricity and density distribution towards the hematoma periphery. The regional TBM model outperformed alternative clinician based and machine-learning methods.

**Conclusions**

TBM discriminated and visualized regional features of hematoma expansion from presentation NCCT scans. This provides preliminary evidence for regional radiographic features of hematoma expansion. It has the potential to improve hematoma expansion prediction and understanding of hematoma expansion pathophysiology.

## Introduction

Within hours after spontaneous intracerebral hemorrhage (ICH) has first been identified on non-invasive brain imaging, hematoma expansion occurs in approximately one quarter to one third of patients.[1-3] It is an independent prognosticator of poor functional and survival outcomes, contributing to mass effect and injury to the surrounding brain tissue.[1,4-6] Consistently identified predictors of hematoma expansion are (1) larger initial hematoma volume and (2) shorter time from symptom onset to the first non-contrast computed tomography (NCCT) scan.[6-10] More recently, the location of the hematoma has been suggested to influence hematoma expansion and final hematoma volume to influence neurological outcome.[5,11-14] The observed prognostic relationships between location and hematoma expansion may be driven by the eloquence and the mechanical properties of the brain tissue surrounding the hematoma.[15] Therefore, the effects of hematoma expansion on outcome are likely to be modified by the initial and final hematoma volumes, the hematoma location, and the time from symptom onset to the initial NCCT scan.

The ability to accurately predict hematoma expansion could become a patient selection tool for development and testing of targeted ICH therapies.[5,16] We previously developed transport-based morphometry (TBM) to quantify and visualize NCCT radiographic markers of hematoma expansion, finding comparable performance to standard clinician-based and machine learning methods.[17] The advantage of this unique approach is that it is generative, permitting visual interpretation of the predictive features identified by the model.[18-21] In the present study, we hypothesized that NCCT features that predict hematoma expansion may be influenced by regional properties of the tissue surrounding the hematoma. Visualization of regional radiographic markers of hematoma expansion may, in turn, provide insight into the underlying biophysical mechanisms for expansion. Our aims were (1) identify quantitative, region-based NCCT predictors of hematoma expansion, (2) visualize and interpret region-based NCCT features of hematoma expansion, and (3) evaluate the performance of a region-based NCCT hematoma expansion prediction model accounting for initial hematoma volume and the time from symptom onset to the first NCCT scan, with comparison to established or emerging alternative methods.

## Methods

### Study Population

This analysis of spontaneous ICH patients was prepared according to the Standards for Reporting Diagnostic accuracy studies (STARD) guideline.[22] Subjects for model derivation were recruited from the Ethnic/Racial Variations of Intracerebral Hemorrhage (ERICH) study, a prospective, case-control, multi-center study of ICH with emphasis on recruitment of balanced proportions of non-Hispanic white, non-Hispanic black and Hispanic ICH cases.[23] Subjects for model external validation were recruited from the Virtual International Trials in Stroke Archive (VISTA), which is an international multi-center collaborative pooled repository of anonymized patient data from randomized clinical trials.[24] Inclusion criteria were: (1) age ≥18 years, (2) presentation with CT-proven ICH, (3) at least one subsequent CT scan at 24±6 hours after the initial scan, (4) available baseline clinical and laboratory data, (5) supratentorial location, and (6) initial ICH volume 2-60mL. The initial volume threshold was set to distinguish from cerebral microbleeds and large initial hematoma volumes, for whom additional expansion would not be considered clinically significant.[11,25] Exclusion criteria were: (1) primary intraventricular hemorrhage (IVH), (2) ICH related to suspected secondary causes, and (3) surgical ICH evacuation. All required Institutional Review Board reviews and approvals were completed at the respective institutions.

### Demographic, clinical and radiographic information

Baseline demographic, clinical and radiographic information collected included age, sex, self-reported race/ethnicity, presentation systolic blood pressure (mmHg), Glasgow Coma Scale score, international normalized ratio, activated partial thromboplastin time (seconds), presence of intraventricular hemorrhage (IVH), and time from symptom onset to first NCCT scan (min). Hematoma growth rate was defined as hematoma volume at 24 ± 6 hours minus initial volume divided by the time from symptom onset to first NCCT (mL/min).

**NCCT scan preprocessing, segmentation and volumetry**

The pre-processing pipeline has been previously described.[17] In brief, (1) the de-identified NCCT scans at presentation and 24±6 hours were transferred in Digital Imaging and Communications in Medicine (DICOM) format to a central workstation; (2) 3D DICOM CT images were converted to a 3-channel NumPy array; (3) windowing was performed and the region corresponding to skull was removed; (4) images were registered to a population-based, high-resolution NCCT template and re-sized to dimensions of 150 x 190 x 120 with voxel spacings of 1 x 1 x 1mm; (5) hematoma regions were segmented using our previously described convolutional neural network; (6) segmented hematoma images located in the left hemisphere were translated across midline; (7) a 3D curvature-driven gaussian filter was applied to smooth the segmented hematoma images; and (8) each segmented hematoma image was normalized so that the sum of its intensities was equal to 1.[26] This protocol was intended to remove differences in CT acquisition methods, facilitating quantitative comparisons of segmented hematoma images. After NCCT scan preprocessing, hematoma volumes at presentation and 24±6 hours were measured by multiplying the number of voxels (volumetric pixels) in the segmented hematoma region by the volume of each voxel (1x1x1 mm).

All pre-processed images and their corresponding segmentations were visually inspected (N.I.,

neurosurgeon, H.K., K.P. medical imaging scientists) to evaluate for skull-stripping and registration errors prior to further analysis. Segmentations were adjudicated at weekly meetings and efforts were made to achieve a consensus in cases of inter-adjudicator differences in opinion. When adjudicators disagreed with the automated method, segmentations were manually edited using the open-source software platform 3D Slicer version 4.10.2 (National Institutes of Health, Bethesda, MD). The code used for preprocessing is available as an open-source package.[27]

**Outcomes**

The thresholds for significant hematoma expansion were defined by region and according to their effect on clinical outcome. This was in accordance with previously described methods.[11] The initial hematoma volume was measured from the first NCCT scan at hospital presentation, and the final hematoma volume was measured at 24±6 hours after the initial NCCT scan. Poor clinical outcome was defined as a modified Rankin Scale score of 3-6 at 90 days. Hematoma expansion was defined as ≥6 mL and/or ≥33% hematoma volume increase from the NCCT scan at initial presentation to 24±6 hours. This was chosen to reflect the threshold most used by established hematoma expansion prediction methods and in ICH clinical outcomes studies.[2,28]

The three pre-specified hematoma regions comprised lobar, basal ganglia, and thalamus. For each region, area under the receiver operating curve (AUROC) analyses assessed the relationship between final hematoma volume and poor clinical outcome. The Youden Index identified the final hematoma volume cutoff for predicting poor functional outcome.[29] Clinically significant hematoma expansion was defined per region as (1) ≥6mL and/or 33% increase in hematoma volume between the presentation and 24±6-hour NCCT scans, and (2) crossing the regional final hematoma volume threshold for poor functional outcome.

### Defining hematoma location

A statistical neuroanatomical segmented MRI atlas was used to identify and outline each of the lobar, basal ganglia and thalamic regions.[30] This was registered to the co-ordinate system of the pre-processed imaging dataset using the Symmetric Normalization method from the Advanced Normalization Tools in Python package version 0.1.8. The Dice Similarity Co-efficient measured the percentage overlap between each segmented hematoma and each of the three regions. The assigned location for each segmented hematoma was defined as whichever of the three regions had the maximal percentage overlap **(Figure 2)**.

### 3D transport-based morphometry

The continuous linear optimal transportation method has been previously described.[17,18,20,31] In brief, we assumed that hematoma expansion occurs as a continuous process of red blood cell movement and that normalized Hounsfield Unit (HU) pixel intensity values represent relative measurements of blood density. A reference image ($I_0$) represented the average hematoma appearance for the dataset. The optimal transport map, a unique linear embedding, was calculated with respect to $I_0$. The minimizer of the optimal transport problem was the squared Wasserstein distance between the densities (images) $\mathrm{I}_0$ and $\mathrm{I}_1$, denoted as $W_2^2(I_0, I_1)$. It had units of *normalized intensity* × *m*$^2$, which quantify the relative movement of blood density from the reference image ($\mathrm{I}_0$ ) to any given hematoma image ($\mathrm{I}_1$).

### Regional hematoma expansion model training

Solving the continuous optimal transportation problem resulted in a 3D mass preserving (MP) map. We obtained the MP map for the segmented presentation NCCT hematoma image of each patient. For each region, these data were separated into independent cohorts for model derivation and validation. There was no crossover between the derivation and validation data.

Because the dimensionality of the hematoma image features in transport space was much higher than the number of data samples, the principal components analysis (PCA) method was used to reduce the dimensionality of the MP maps and retain the top $n$ directions that explained 95% of the variance in the data.[18,19,31] Initial hematoma volume and time from symptom onset to the first NCCT scan are established modifiers of hematoma expansion.[8,10,32,33] To account for these variables, initial hematoma volume and hematoma growth rate were concatenated with the principal components of the image feature matrix. Our previously described penalized linear discriminant analysis (pLDA) method was used to assess the relationship between this multivariable feature matrix and 24-hour hematoma expansion.[17,34] For each hematoma region, we retained the top pLDA direction $w^0$ that maximally separated the expansion and no expansion groups to train a binary classifier. The code used for model derivation is available as an open-source package.[27]

**Regional hematoma expansion model testing**

For each region, the multivariable feature matrix of the validation dataset was projected onto $w^o$ and area under the receiver operator characteristic curve (AUROC) analyses evaluated the model's performance. The Youden Index identified the optimal threshold for classification in the derivation dataset.[29] This was applied to the validation dataset and classification accuracy was assessed using sensitivity, specificity and balanced accuracy scores.

An overall TBM model, weighted by the proportion of patients in each region, was derived in validation dataset by averaging the weighted predicted probabilities for each of the three regional TBM models. Mean AUROC analyses evaluated the overall model's performance. Classification accuracy was assessed using mean sensitivity, specificity and balanced accuracy scores.

**Visualization**

To visualize the presentation NCCT hematoma image features discriminating hematoma expansion for each region, we inverted the projections of the derivation dataset onto the principal pLDA direction $w^0$:

$$w_{pLDA} = \overline{X}_{train} + \sigma w^0$$

where $w_{pLDA}$ is the direction and magnitude of hematoma feature difference between the expansion and no expansion groups, $\overline{X}_{train}$ is the center of $X_{train}$ and $\sigma$ has units of standard deviations of the data projected along $w^0$.

**Feature-based model**

The inverse transformations of the TBM model permitted visualization of regional NCCT features of hematoma expansion. The discriminating features for each region were then quantified as functions of the image properties of the registered and normalized segmented hematomas at presentation. In the derivation dataset, a linear classifier was trained on each of the resultant regional image feature matrices. Its performance in the validation dataset was assessed using AUROC analyses. The Youden Index identified the optimal threshold for classification in the derivation dataset.[29] This was applied to the validation dataset and classification accuracy was assessed using sensitivity, specificity and balanced accuracy scores.

An overall feature-based model, weighted by the proportion of patients in each region, was derived by averaging the predicted probabilities for each of the three feature-based regional models in the validation dataset. Mean AUROC analyses evaluated the overall feature-based model's performance. Classification accuracy was assessed using mean sensitivity, specificity and balanced accuracy scores.

**Comparison to alternate clinician-based and machine learning methods**

The existing BAT, BRAIN, HEAVN, NAG and 10-point clinical hematoma expansion prediction scores were computed by two independent trained raters who were blinded to outcomes information (A.B-G, neurosurgeon in-training; K.E.N., neurosurgeon in-training) and their performance in the validation dataset was compared to the regional TBM and feature-based models using AUROC analyses.[8-10,32,35]

Previously described machine learning models for NCCT-based classification of hematoma expansion including support vector machine (SVM), K-nearest neighbors (KNN), logistic regression, random forest, 3D ResNet 18 and 3D DenseNet 121 in were trained on the registered and normalized segmented hematoma images from the derivation dataset. Their classification performance was assessed on the registered and normalized segmented presentation NCCT hematoma images in the validation dataset. This was compared to the regional TBM model using AUROC analyses.

**Statistical analyses**

Statistical analyses were performed using Python version 3.12.0 and Matlab R2022a. Baseline demographic, clinical and radiographic parameters were compared between the derivation and validation datasets using Pearson's $\chi^2$ square or Fisher's exact test for categorical data and student's *t*-test or Mann-Whitney U test for continuous data, as appropriate. Statistical significance was defined as $p<0.05$. Missing data was not imputed.

## Results

**Dataset composition**

Of the 1,148 spontaneous ICH patients in the ERICH repository with available presentation and 24±6 hr NCCT scans, 574 were excluded (126 for corrupted DICOM files, 156 for infratentorial ICH location, 57 for surgical ICH evacuation, 50 for primary IVH, 2 for multicompartmental or bilateral ICH, 42 for registration errors, 19 for segmentation errors and 122 for initial ICH volume <2 or >60mL) **(Figure 1)**. The derivation dataset comprised the remaining 574 (mean (SD) age 61.47 (13.54) years; 34.2% (n=193) female), 32.6% (n=184) Black, 36.1% (n=204) Hispanic and 31.2% (n=176) White patients. In the derivation dataset, the mean (SD) and median [IQR] hematoma volumes at presentation were 19.46 (14.95) and 14 [7-29] mL, respectively. The mean (SD) and median [IQR] hematoma volumes at 24±6 hours were 23.13 (19.29) and 16 [8-33] mL, respectively. Hematoma expansion was present in 17.1% (n=98) patients.

Of the 262 spontaneous ICH patients in the VISTA repository with available presentation and 24±6 hour interval NCCT scans, 92 were excluded (28 for corrupted DICOM files, 9 for infratentorial ICH location, 3 for surgical ICH evacuation and 26 for initial ICH volume <2 or >60mL) **(Figure 1)**. The validation dataset comprised the remaining 196 patients (mean (SD) age 64.83 (12.45) years; 35.0% (n=70) female). In the validation dataset, the mean (SD) and median [IQR] hematoma volumes at presentation were 20.56 (15.33) and 15 [7-31] mL, respectively. The mean (SD) and median [IQR] hematoma volumes at 24±6 hours were 26.33 (25.62) and 19 [8-36] mL, respectively. Hematoma expansion was present in 24.0% (n=47) patients. Comparisons of baseline demographic and clinical characteristics between the expansion and no expansion groups for each of the derivation and validation datasets are presented in **Supplementary Table 1.** There were between dataset differences in mean age (derivation: 61.5±13.5 vs. validation: 64.8±12.5 years, p=0.002), admission systolic blood pressure (derivation: 187±36 vs. validation: 181±28mmHg, p=0.048), time from symptom onset to first NCCT scan (derivation: 971±1569 vs. validation: 109±39 minutes, p<0.001), hematoma growth rate (derivation: 0.12±0.23 vs. validation: 0.22±0.20 mL/minute, p<0.001) and presence of intraventricular hemorrhage (derivation: 42.7% vs. validation: 30.4%, p=0.003). Clinical information was missing for 10 patients in the derivation dataset and 0 patients in the validation dataset.

**Region categories**

Of the 574 patients in the derivation dataset, 162 (28.2%) were categorized as lobar, 257 (44.8%) basal ganglia, and 155 (27.0%) thalamic hematomas. In the derivation dataset, hematoma expansion was present in 22.2% (n=36) patients in the lobar, 19.1% (n=49) in the basal ganglia and 8.4% (n=13) in the thalamic regions. Of the 196 patients in the validation dataset, 31 (15.8%) were categorized as lobar, 119 (60.7%) were basal ganglia and 46 (23.5%) were thalamic hematomas. In the validation dataset, hematoma expansion was present in 29.0% (n=31) patients in the lobar, 27.8% (n=33) in the basal ganglia and 10.9% (n=5) in the thalamic regions. For each region, visual comparisons of segmented hematoma images did not identify a discernible difference between the hematoma expansion and no expansion groups **(Figure 2)**.

**Regional definitions of hematoma expansion**

The optimal hematoma volume cut point at 24±6 hours predictive of 90-day mRS 3-6 was 23mL for the lobar region, 22mL for the basal ganglia region and 13mL for the thalamic region. For the lobar region, this was associated with an AUROC of 0.780, sensitivity of 88.0% and specificity of 68.0%. For the basal ganglia region, this was associated with an AUROC of 0.710, sensitivity of 61.0% and specificity of 82.0%. For the thalamic region, this was associated with an AUROC of 0.680, sensitivity of 52.0% and specificity of 68.0%.

**Regional TBM model discriminated 24-hour hematoma expansion from NCCT**

For each regional cohort, the most discriminant direction in transport space $w^0$ was identified in the derivation dataset. The validation data projected on to $w^0$ demonstrated a separation of the mean probability distributions for each of the lobar, thalamus and basal ganglia regions. These were statistically significant. **(Figure 3)**.

***Lobar***

In the validation dataset, the pLDA classifier trained on $w^0$ predicted expansion with an area under the receiver operating curve (AUROC) of 0.773 with initial hematoma volume and hematoma growth rate included in the model. The balanced accuracy, sensitivity and specificity were 72.7%, 100.0% and 45.5%, respectively, for the lobar TBM model. **(Figure 4)**.

***Basal Ganglia***

In the validation dataset, the pLDA classifier trained on $w^0$ predicted expansion with an area under the receiver operating curve (AUROC) of 0.739 with initial hematoma volume and hematoma growth rate included in the model. The balanced accuracy, sensitivity and specificity were 68.2%, 81.8% and 54.6%, respectively, for the basal ganglia TBM model. **(Figure 4)**.

***Thalamus***

In the validation dataset, the pLDA classifier trained on $w^0$ predicted expansion with an area under the receiver operating curve (AUROC) of 0.820 with initial hematoma volume and hematoma growth rate included in the model. The balanced accuracy, sensitivity and specificity were 63.4%, 100.0% and 26.8%, respectively, for the thalamus TBM model. **(Figure 4)**.

***Combined model***

In the validation dataset, the pLDA classifier trained on $\mathrm{w}^0$ predicted expansion with a mean area under the receiver operating curve (AUROC) of 0.787 with initial hematoma volume and hematoma growth rate included in the model. The balanced accuracy, sensitivity and specificity were 67.2%, 88.2% and 47.6%, respectively, for the combined regional TBM model. **(Figure 4)**.

**TBM discovered interpretable NCCT features of 24-hour hematoma expansion**

TBM-generated images identified regional NCCT features discriminating hematoma expansion.

For each region, the inverse transformations of the hematoma features projected onto $w^0$ were plotted in units of standard deviations (SD) of the pixel intensity distribution along $w^0$. For the lobar region, the features associated with expansion were increasing size, peripheral density distribution, and density heterogeneity. For the basal ganglia region, the features associated with expansion were increasing size, shape eccentricity, peripheral density distribution and density heterogeneity. For the thalamic region, the features associated with expansion were increasing size, shape eccentricity and peripheral density distribution. The inverse transformations and their associated probability density distributions discriminating the expansion versus no expansion groups for each region are presented in **Figure 3**. The inverse transformations overlaid onto NCCT images, highlighting examples of features associated with expansion for each region, are presented in **Figure 5**.

**Regional feature-based model discriminated 24-hour hematoma expansion from NCCT**

The principal features identified from the inverse transformations of the TBM model were quantified from the registered, segmented and normalized NCCT hematoma images in the native image domain. The resultant multivariable image feature matrix comprised NCCT features and hematoma growth rate.

***Lobar***

In the validation dataset, the lobar feature-based model predicted expansion with an area under the receiver operating curve (AUROC) of 0.808. The balanced accuracy, sensitivity and specificity were 67.7%, 64.1% and 72.7%, respectively, for the lobar feature-based model. **(Supplementary Figure 1)**.

### *Basal Ganglia*

In the validation dataset, the basal ganglia feature-based model predicted expansion with an area under the receiver operating curve (AUROC) of 0.767. The balanced accuracy, sensitivity and specificity were 71.0%, 90.9% and 51.2%, respectively, for the basal ganglia feature-based model. **(Supplementary Figure 1)**.

### *Thalamus*

In the validation dataset, the thalamus feature-based model predicted expansion with an area under the receiver operating curve (AUROC) of 0.781. The balanced accuracy, sensitivity and specificity were 70.5%, 80.0% and 61.0%, respectively, for the thalamus feature-based model. **(Supplementary Figure 1)**.

### *Combined model*

In the validation dataset, the combined regional feature-based model predicted expansion with a mean area under the receiver operating curve (AUROC) of 0.775. The balanced accuracy, sensitivity and specificity were 67.1%, 71.7% and 62.5%, respectively, for the combined regional feature-based model. **(Supplementary Figure 1)**.

## Predictive performance of region-based models with comparison to alternate clinician-based, and machine learning methods

Clinician-based scores for hematoma expansion prediction were assessed in the validation dataset.

The AUROC for predicting 24-hour hematoma expansion was 0.693 for BAT, 0.709 for BRAIN, 0.713 for HEAVN, 0.609 for NAG, and 0.603 for the 10-point score. Both the TBM and the feature-based models demonstrated superior performance to clinician-based methods for predicting 24-hour hematoma expansion in the validation dataset. **(Figure 6)**.

Alternate machine and deep learning methods for hematoma expansion prediction were trained using the training dataset and assessed in the test dataset. The AUROC for predicting 24-hour hematoma expansion was 0.665 for SVM, 0.656 for KNN, 0.481 for logistic regression, 0.739 for random forest, 0.728 for ResNet 18 and 0.723 for DenseNet 121 in the validation dataset. The random forest and neural networks classifiers achieved high specificity values of 98.7% for random forest, 98.0% for ResNet 18 and 87.2% for DenseNet121. However, they demonstrated low sensitivity values of 0% for random forest, 0% for ResNet 18 and 25.5% for DenseNet 121 in the validation dataset. Not only did the TBM and feature-based models demonstrate improved overall classification performance, but they also demonstrated improved sensitivity over alternate machine learning and deep learning methods for predicting 24-hour hematoma expansion in the test dataset **(Supplementary Figure 2).**

## Discussion

Location influences the relationship between initial hematoma volume and clinical outcome.[14,15,36] Location-specific initial hematoma volume thresholds have been found to predict functional prognosis in spontaneous ICH patients.[15] Interactions between final hematoma volume, hematoma expansion and location have also been identified.[11,13] A proposed critical regional tolerance threshold suggests that hematoma expansion may influence outcome only if the final volume exceeds a certain threshold for a given location.[11] These consistent observations indicate that the clinical significance of hematoma volume and hematoma expansion are likely to be modulated by the location of the hematoma. In this study, we hypothesized that NCCT biomarkers of hematoma expansion may also be affected by location. Using data from the ERICH natural history study for algorithm derivation and the VISTA clinical trials database for algorithm validation, we developed quantitative NCCT radiomics models for supratentorial, spontaneous ICH patients that were stratified into lobar, basal ganglia, and thalamic regions. Expansion was defined if regional 24±6hr volume thresholds for predicting 90-day poor clinical outcome were met in addition to a ≥6mL and/or 33% hematoma volume increase from baseline to 24±6 hours. We applied our previously described transport-based morphometry (TBM) framework to model the relative pixel intensities of segmented NCCT hematoma images as surrogates for relative measures of blood density.[17] For identifying NCCT features discriminating hematoma expansion, this approach achieved a statistically significant separation of the mean probability distributions between the expansion and no expansion groups for each location. In addition, we found regional differences in the proportion of patients experiencing expansion, with thalamic hematomas being less likely to expand than their basal ganglia or lobar counterparts. These regional differences suggest unique biophysical mechanisms for hematoma expansion.

,

We previously published a TBM method identifying NCCT features of hematoma expansion and in the present study, we expanded upon this approach by determining regional morphological features of expansion.[17] Classifying hematomas by region achieved an overall AUROC of 0.79 versus 0.71, when compared to our previous whole brain model. This improved performance may have, in part, resulted from the utilization of location-specific definitions for hematoma expansion that account for the clinical importance of final hematoma volume as well as relative or absolute change in volume over time. An advantage of the TBM method is its interpretability. Inverting the transport representations of each regional TBM model provided radiographic insight into the image features used for its predictions. These visual differences between the image features discriminating expansion suggest evidence for regional differences in the underlying processes driving hematoma expansion.[14] For example, shape eccentricity may represent blood cell movement via the path of least resistance, corresponding to fiber or perivascular space orientations within the brain parenchyma.[37] In the basal ganglia region, this mode of expansion was predominantly oriented towards the lateral ventricle. However, it is important to note that these observations are speculative and real associations with biophysical phenomena cannot be concluded. The quantitative nature of our method also overcame the subjectivity inherent to the qualitative NCCT features described by clinicians, which do not account for location. The observed regional differences among NCCT morphometric features of expansion and improvements to TBM's classification performance support our hypothesis that region-specific modeling may provide a more accurate approach to predicting hematoma expansion.

The image features identified by each regional TBM model were quantified in the native image domain. These were used to derive a regional feature-based model of hematoma expansion which had comparable performance to the TBM model. The advantage of the feature-based model was its explainability and simplicity, requiring fewer dimensions to achieve a similar classification performance. Additionally, the consistency between the features identified by the TBM model that were corroborated in the feature-based model, and those previously described by clinicians, lends support to the validity of our quantitative approach.[38,39] In the validation dataset, the overall combined AUROC for the TBM model was 0.787 versus 0.775 for the feature-based model. Comparisons of classification accuracy found higher sensitivity with the TBM model, whereas the feature-based model had higher specificity. Compared to our previously published study, we have expanded the derivation dataset and demonstrate generalizability between derivation natural history and validation clinical trials datasets. There was no crossover between the model parameters from our previous publication. Our study contributions are (1) identification of quantitative, regional NCCT signatures for hematoma expansion, (2) application of regional definitions for hematoma expansion prediction, and (3) development and validation of an automated, regional NCCT hematoma expansion prediction model across heterogeneous ICH patient cohorts. With further optimization, regional quantitative NCCT feature identification may permit improved patient selection for clinical trials targeting ICH expansion with hemostatic therapies or surgical evacuation. However, for effective development of future therapies, the optimal definition of hematoma expansion that is associated with a reversible change in clinical outcome requires further investigation.

The results of this study are preclinical and hypothesis-generating in nature. While they do not represent a readily implementable tool for use in clinical practice, they suggest that mathematical modeling of NCCT image features could become a viable method for risk-stratification and identification of therapeutic targets in spontaneous intracerebral hemorrhage patients. Prioritizing interpretability will be crucial to generalizability and could lead to an improved understanding of the physiology of the disease. Initial hematoma volume, a core component of four of the five clinician-based scores, was also observed to be dominant predictor of hematoma expansion in our models.[8-10,32] Future studies, which are currently underway, will seek to incorporate demographic and clinical information and effectively weight the relative predictive features in the models. NCCT is the most frequently obtained imaging study in ICH patients and represents a practical modality for expansion prediction. However, as demonstrated by the limited resolution of the TBM inverse transformations **(Figures 5 and 6)**, the lack of pixel density information on NCCT may incompletely distinguish features of expansion. Recent promising results of the CTA spot sign for predicting hematoma expansion and response to hemostatic therapies suggests the potential for combined NCCT and CTA models to significantly improve classification performance.[16] Intraventricular hemorrhage is an additional image feature not addressed in the present study that is likely to modulate hematoma expansion and warrants further analysis.[33] Increasing the number of region categories and size of the datasets used for training and validation will also be necessary to account for the heterogeneity of the spontaneous ICH population.

There are several important limitations to this study which must be acknowledged. Firstly, our model is limited by its retrospective design. Although we included representative multi-national populations of patients enrolled in ICH studies and an external validation cohort, our models lack prospective validation in a real-world setting. Therefore, their potential applicability as clinical tools is undetermined. The baseline differences between the derivation and validation cohorts, particularly in the time from symptom onset to first NCCT, suggest real-world differences between natural history and clinical trials data. As these differences were not adjusted for, our results are subject to potential confounding bias. This should be addressed in future studies using methods such as case-matched patient sampling or model adjustment for confounders. The model performance is also subject to the reliability of the pre-processing and segmentation methods, which required manual editing in a proportion of cases. This has limited effective automation of our approach and requires further tuning in future studies. Regions were defined from a registered, segmented MRI atlas and are limited by a lack of validation from expert radiologists. For hematomas that crossed locations, the greatest proportion of overlap was used to define the dominant region. In future larger studies, subgroup analyses by hematomas which crossed location and by hematoma volume would help to elucidate their relative effects on prediction modeling. Model development was affected by the imbalanced nature of the dataset with small numbers of positive cases in certain locations. The optimal approach for dealing with class imbalance requires further investigation in future studies. Despite reasonable AUROCs, our model performance demonstrated high sensitivity and relatively low specificity, thereby limiting its accuracy. Although our model outperformed alternate methods, future performance optimization should focus on larger, multimodal datasets combining NCCT, CTA and clinical features. Clinician-based scores were performed by two raters. In the absence of assessments of inter-rater reliability and averaged results across multiple raters, they may be subject to rater bias. Finally, because this study was not powered, the validation sample size required to establish predictive performance remains undefined.

## Conclusions

Region-based modeling of non-contrast computed tomography (NCCT) in spontaneous intracerebral hemorrhage (ICH) patients attests to regional radiographic features of hematoma expansion. Transport-based morphometry (TBM) discriminated and visualized regional features of 24-hour hematoma expansion from presentation NCCT scans. A combined regional TBM model, comprising lobar basal ganglia and thalamic regions, achieved an AUROC of 0.79 in the validation dataset. The information used by the model was reproducible using feature-based statistical analysis methods. It outperformed current clinician-based methods. This quantitative and interpretable approach has the potential to improve hematoma expansion prediction and understanding of hematoma expansion pathophysiology.

**Appendix 2**

**Co-investigators: VISTA-ICH Steering Committee**

| **Name** |
|---|
| DF Hanley |
| K Butcher |
| S Davis |
| B Gregson |
| KR Lees |
| P Lyden |
| S Mayer |
| K Muir |
| E Sandset |
| T Steiner |

## References

1. Brouwers HB, Greenberg SM. Hematoma expansion following acute intracerebral hemorrhage. *Cerebrovasc Dis*. 2013;35(3):195-201. doi:10.1159/000346599
2. Dowlatshahi D, Demchuk AM, Flaherty ML, Ali M, Lyden PL, Smith EE. Defining hematoma expansion in intracerebral hemorrhage: relationship with patient outcomes. *Neurology*. Apr 5 2011;76(14):1238-44. doi:10.1212/WNL.0b013e3182143317
3. Hemphill JC, 3rd. Hematoma Expansion in ICH: Targeting Epidemiology or Biology? *Neurocrit Care*. Aug 2019;31(1):9-10. doi:10.1007/s12028-019-00752-1
4. Huttner HB, Gerner ST, Kuramatsu JB, et al. Hematoma Expansion and Clinical Outcomes in Patients With Factor-Xa Inhibitor-Related Atraumatic Intracerebral Hemorrhage Treated Within the ANNEXA-4 Trial Versus Real-World Usual Care. *Stroke*. Feb 2022;53(2):532-543. doi:10.1161/strokeaha.121.034572
5. Kuohn LR, Witsch J, Steiner T, et al. Early deterioration, hematoma expansion, and outcomes in deep versus lobar intracerebral hemorrhage: the FAST trial. *Stroke*. 2022;53(8):2441-2448.
6. Morotti A, Boulouis G, Charidimou A, et al. Hematoma Expansion in Intracerebral Hemorrhage With Unclear Onset. *Neurology*. May 11 2021;96(19):e2363-e2371. doi:10.1212/wnl.0000000000011895
7. Morotti A, Dowlatshahi D, Boulouis G, et al. Predicting Intracerebral Hemorrhage Expansion With Noncontrast Computed Tomography: The BAT Score. *Stroke*. May 2018;49(5):1163-1169. doi:10.1161/strokeaha.117.020138

8. Wang X, Arima H, Al-Shahi Salman R, et al. Clinical prediction algorithm (BRAIN) to determine risk of hematoma growth in acute intracerebral hemorrhage. *Stroke*. Feb 2015;46(2):376-81. doi:10.1161/strokeaha.114.006910
9. Miyahara M, Noda R, Yamaguchi S, et al. New Prediction Score for Hematoma Expansion and Neurological Deterioration after Spontaneous Intracerebral Hemorrhage: A Hospital-Based Retrospective Cohort Study. *J Stroke Cerebrovasc Dis*. Sep 2018;27(9):2543-2550. doi:10.1016/j.jstrokecerebrovasdis.2018.05.018
10. Brouwers HB, Chang Y, Falcone GJ, et al. Predicting hematoma expansion after primary intracerebral hemorrhage. *JAMA Neurol*. Feb 2014;71(2):158-64. doi:10.1001/jamaneurol.2013.5433
11. Morotti A, Li Q, Nawabi J, et al. Volume Tolerance and Prognostic Impact of Hematoma Expansion in Deep and Lobar Intracerebral Hemorrhage. *Stroke*. May 2025;56(5):1224-1231. doi:10.1161/strokeaha.124.049008
12. Delcourt C, Huang Y, Arima H, et al. Hematoma growth and outcomes in intracerebral hemorrhage: the INTERACT1 study. *Neurology*. Jul 24 2012;79(4):314-9. doi:10.1212/WNL.0b013e318260cbba
13. Song L, Qiu X-M, Guo T-T, et al. Association between anatomical location and hematoma expansion in deep intracerebral hemorrhage. *Frontiers in Neurology*. 2022;12:749931.
14. Roh D, Sun C-H, Murthy S, et al. Hematoma expansion differences in lobar and deep primary intracerebral hemorrhage. *Neurocritical care*. 2019;31:40-45.
15. Ironside N, Chen CJ, Dreyer V, Christophe B, Buell TJ, Connolly ES. Location-specific differences in hematoma volume predict outcomes in patients with spontaneous intracerebral hemorrhage. *Int J Stroke*. Jan 2020;15(1):90-102. doi:10.1177/1747493019830589
16. Broderick JP, Naidech AM, Elm JJ, et al. Recombinant factor VIIa versus placebo for spontaneous intracerebral haemorrhage within 2 h of symptom onset (FASTEST): a multicentre, double-blind, randomised, placebo-controlled, phase 3 trial. *Lancet*. Feb 21 2026;407(10530):773-783. doi:10.1016/s0140-6736(26)00097-8
17. Ironside N, El Naamani K, Rizvi T, et al. Predictive modeling of hematoma expansion from non-contrast computed tomography in spontaneous intracerebral hemorrhage patients. *Elife*. Dec 23 2025;14doi:10.7554/eLife.105782
18. Wang W, Slepčev D, Basu S, Ozolek JA, Rohde GK. A linear optimal transportation framework for quantifying and visualizing variations in sets of images. *Int J Comput Vis*. Jan 1 2013;101(2):254-269. doi:10.1007/s11263-012-0566-z
19. Kundu S, Ashinsky BG, Bouhrara M, et al. Enabling early detection of osteoarthritis from presymptomatic cartilage texture maps via transport-based learning. *Proc Natl Acad Sci U S A*. Oct 6 2020;117(40):24709-24719. doi:10.1073/pnas.1917405117
20. Kundu S, Kolouri S, Erickson KI, Kramer AF, McAuley E, Rohde GK. Discovery and visualization of structural biomarkers from MRI using transport-based morphometry. *NeuroImage*. 2018;167:256-275.
21. Kolouri S, Park S, Thorpe M, Slepčev D, Rohde GK. Optimal Mass Transport: Signal processing and machine-learning applications. *IEEE Signal Process Mag*. Jul 2017;34(4):43-59. doi:10.1109/msp.2017.2695801
22. Bossuyt PM, Reitsma JB, Bruns DE, et al. STARD 2015: an updated list of essential items for reporting diagnostic accuracy studies. *Bmj*. Oct 28 2015;351:h5527. doi:10.1136/bmj.h5527
23. Woo D, Rosand J, Kidwell C, et al. The Ethnic/Racial Variations of Intracerebral Hemorrhage (ERICH) study protocol. *Stroke*. Oct 2013;44(10):e120-5. doi:10.1161/strokeaha.113.002332
24. Ali M, Bath P, Brady M, et al. Development, expansion, and use of a stroke clinical trials resource for novel exploratory analyses. *Int J Stroke*. Feb 2012;7(2):133-8. doi:10.1111/j.1747-4949.2011.00735.x
25. Morotti A, Boulouis G, Dowlatshahi D, et al. Standards for Detecting, Interpreting, and Reporting Noncontrast Computed Tomographic Markers of Intracerebral Hemorrhage Expansion. *Ann Neurol*. Oct 2019;86(4):480-492. doi:10.1002/ana.25563
26. Ironside N, Chen CJ, Mutasa S, et al. Fully Automated Segmentation Algorithm for Hematoma Volumetric Analysis in Spontaneous Intracerebral Hemorrhage. *Stroke*. Dec 2019;50(12):3416-3423. doi:10.1161/strokeaha.119.026561
27. Rohde GK. PyTransKit. https://github.com/rohdelab/PyTransKit
28. Mayer SA, Frontera JA, Jankowitz B, et al. Recommended Primary Outcomes for Clinical Trials Evaluating Hemostatic Agents in Patients With Intracranial Hemorrhage: A Consensus Statement. *JAMA Netw Open*. Sep 1 2021;4(9):e2123629. doi:10.1001/jamanetworkopen.2021.23629
29. Ruopp MD, Perkins NJ, Whitcomb BW, Schisterman EF. Youden Index and optimal cut-point estimated from observations affected by a lower limit of detection. *Biom J*. Jun 2008;50(3):419-30. doi:10.1002/bimj.200710415
30. Evans AC, Collins DL, Mills SR, Brown ED, Kelly RL, Peters TM. 3D statistical neuroanatomical models from 305 MRI volumes. *1993 IEEE Conference Record Nuclear Science Symposium and Medical Imaging*

*Conference*. 1993:1813-1817 vol.3.
31. Kolouri S, Tosun AB, Ozolek JA, Rohde GK. A continuous linear optimal transport approach for pattern analysis in image datasets. *Pattern Recognit*. Mar 1 2016;51:453-462. doi:10.1016/j.patcog.2015.09.019
32. Fu J, Hu S, Yang M, et al. A Novel 10-Point Score System to Predict Early Hematoma Growth in Patients With Spontaneous Intracerebral Hemorrhage. *Front Neurol*. 2019;10:1417. doi:10.3389/fneur.2019.01417
33. Morotti A, Boulouis G, Dowlatshahi D, et al. Intracerebral haemorrhage expansion: definitions, predictors, and prevention. *Lancet Neurol*. Feb 2023;22(2):159-171. doi:10.1016/s1474-4422(22)00338-6
34. Basu S, Kolouri S, Rohde GK. Detecting and visualizing cell phenotype differences from microscopy images using transport-based morphometry. *Proc Natl Acad Sci U S A*. Mar 4 2014;111(9):3448-53. doi:10.1073/pnas.1319779111
35. Sakuta K, Sato T, Komatsu T, et al. The NAG scale: Noble Predictive Scale for Hematoma Expansion in Intracerebral Hemorrhage. *J Stroke Cerebrovasc Dis*. Oct 2018;27(10):2606-2612. doi:10.1016/j.jstrokecerebrovasdis.2018.05.020
36. Brissette V, Ouyang M, Yogendrakumar V, et al. Location-Specific Hematoma Volume Tolerance for Spontaneous Intracerebral Hemorrhage: A Pooled Analysis of INTERACT2 and ATACH II. *J Am Heart Assoc*. May 14 2026:e049096. doi:10.1161/jaha.125.049096
37. Bae SB, Kim S. Imaging features of hyperacute intracerebral hemorrhage on multiple MRI sequences within 1 minute from onset during MRI examination: A case report. *Medicine (Baltimore)*. Mar 31 2023;102(13):e33350. doi:10.1097/md.0000000000033350
38. Morotti A, Boulouis G, Charidimou A, et al. Imaging markers of intracerebral hemorrhage expansion in patients with unclear symptom onset. *Int J Stroke*. Oct 2022;17(9):1013-1020. doi:10.1177/17474930211068662
39. Morotti A, Arba F, Boulouis G, Charidimou A. Noncontrast CT markers of intracerebral hemorrhage expansion and poor outcome: A meta-analysis. *Neurology*. Oct 6 2020;95(14):632-643. doi:10.1212/wnl.0000000000010660

## Figures

### Figure 1

Flow diagram illustrating the patient selection process. Abbreviations: NCCT = Non-contrast computed tomography, DICOM = digital imaging and communications in medicine, ICH = intracerebral hemorrhage.

### Figure 2

Example of the preprocessing and region allocation protocol. **(A)** All NCCT scans were skull stripped and registered to a population-based high resolution NCCT template with dimensions of 256 x 256 x 256 and voxel spacings of 1 x 1 x 1mm. A statistical neuroanatomical segmented MRI atlas, registered to the population-based NCCT template, was used to outline the lobar, basal ganglia and thalamic regions. Segmented hematoma images were assigned to a region based upon their maximal percentage overlap with one of the three segmented locations. Examples of segmented presentation NCCT hematoma images separated into groups of hematoma expansion and no hematoma expansion demonstrate a lack of visually discernible difference between the two groups. Abbreviations: NCCT = non-contrast computed tomography.

### Figure 3

Mean probability distributions of the hematoma image features in the test dataset projected onto the most discriminant direction $w_0$ in transport space show the degree of separation between the expansion (red) and no expansion (blue) groups by the learned pLDA classifier boundary for the **(A)** lobar, **(B)** basal ganglia and **(C)** thalamus regional models. Inverse transformations of two two-dimensional axial slice examples of the hematoma morphometric features found by the model to be associated with increasing likelihood of expansion, shown from left to right of the *x*-axis for the **(A)** lobar, **(B)** basal ganglia and **(C)** thalamus regional models.

### Figure 4

Results of the TBM model adjusted for initial hematoma volume and hematoma growth rate in predicting 24-hour hematoma expansion from the validation dataset. AUROC analyses of the performance of the pLDA classifier in the validation dataset for the **(A)** lobar, **(B)** basal ganglia, **(C)** thalamus and **(D)** overall regional models. Abbreviations: TBM = transport-based morphometry, AUROC = Area Under the Receiver Operator Curve, AUC = Area Under the Curve, ROC = Receiver Operator Curve, pLDA = penalized linear discriminant analysis

**Figure 5**

Inverse transformations of the hematoma morphometric features overlaid onto the axial NCCT scan *least* associated with expansion, left, and *most* associated with expansion, right for the **(A)** lobar, **(B)** basal ganglia and **(C)** thalamic TBM models.

**Figure 6**

Performance comparisons between existing NCCT hematoma expansion prediction scores, the overall regional TBM model and the overall regional feature-based model in the validation dataset. Abbreviations: AUROC = Area Under the Receiver Operator Curve, ROC = Receiver Operator Curve, AUC = Area Under the Curve, TBM = transport-based morphometry, NCCT = non-contrast computed tomography, Heavn = Heavn score, Brain = Brain score, NAG = NAG scale, PT = 10-point score, BAT = BAT score, TBM = transport-based morphometry.